\documentclass[10pt,twocolumn,letterpaper]{article}

\usepackage[pagenumbers]{cvpr} %
\usepackage[ruled,vlined]{algorithm2e}  %

\definecolor{cvprblue}{rgb}{0.21,0.49,0.74}
\usepackage{graphicx}
\usepackage[pagebackref,breaklinks,colorlinks,allcolors=cvprblue]{hyperref}
\usepackage{multirow}

\def\methodname{MotionFlow}

\title{\methodname: Attention-Driven Motion Transfer in Video Diffusion Models}

\author{Tuna Han Salih Meral \quad Hidir Yesiltepe  \quad Connor Dunlop \quad Pinar Yanardag \\[2mm]
Virginia Tech\\
{\tt\small \href{https://motionflow-diffusion.github.io}{https://motionflow-diffusion.github.io}}
}

\begin{document}

\twocolumn[{
\maketitle
\begin{center}
    \captionsetup{type=figure}
    \vspace{-1em}
\newcommand{\imwidth}{1\textwidth}

\begin{tabular}{@{}c@{}}

\parbox{\imwidth}{\includegraphics[width=\imwidth]{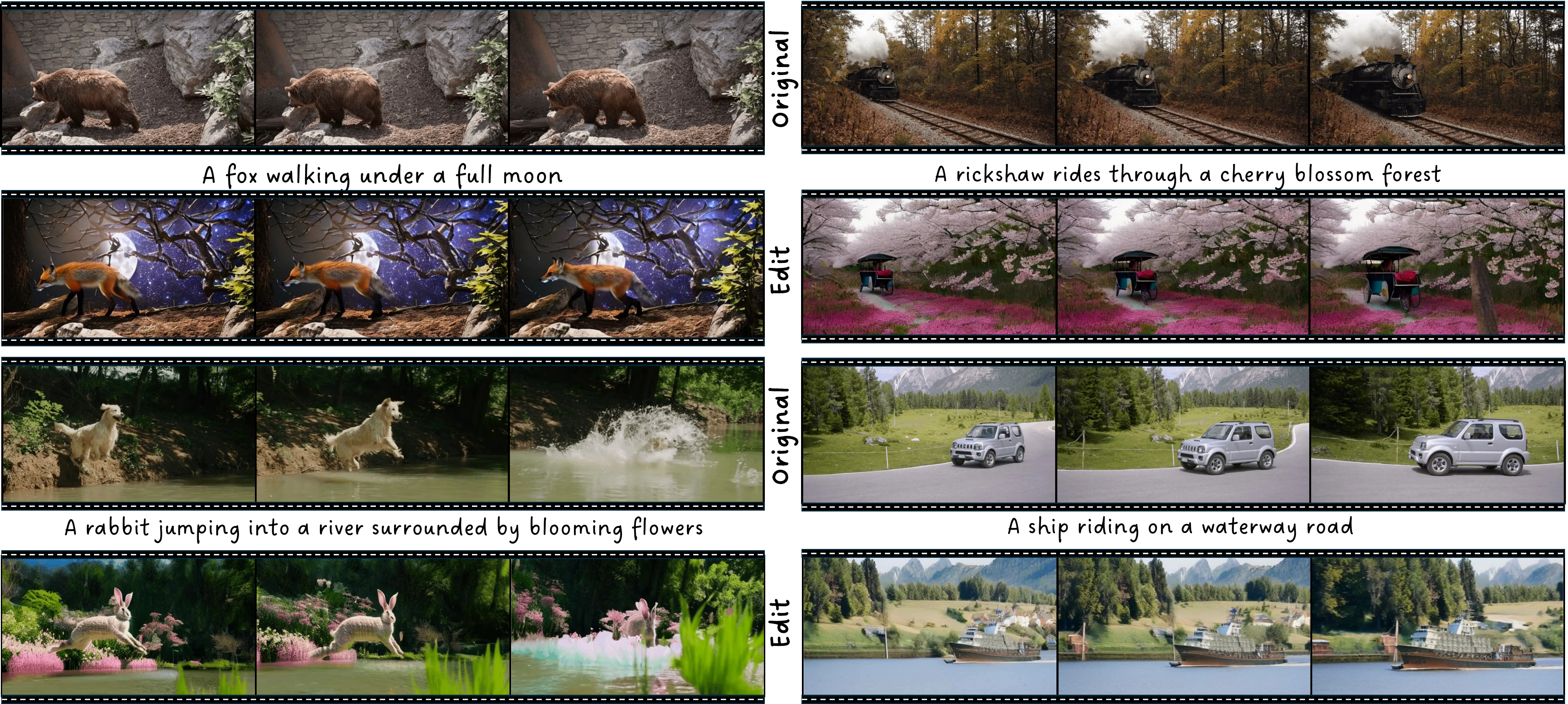}}
\\

\vspace{1em}
\end{tabular}
    \vspace{-2.5em}
    \captionof{figure}{\methodname{} is a training-free method that leverages attention  for motion transfer. Our method can successfully transfer a wide variety of motion types, ranging from simple to complex motion patterns. } 
    \label{fig:teaser}
\end{center}
}] 

\maketitle
\begin{abstract} 

Text-to-video models have demonstrated impressive capabilities in producing diverse and captivating video content, showcasing a notable advancement in generative AI. However, these models generally lack fine-grained control over motion patterns, limiting their practical applicability. We introduce \methodname{}, a novel framework designed for motion transfer in video diffusion models. Our method utilizes cross-attention maps to accurately capture and manipulate spatial and temporal dynamics, enabling seamless motion transfers across various contexts. Our approach does not require training and works on test-time by leveraging the inherent capabilities of pre-trained video diffusion models.  In contrast to traditional approaches, which struggle with comprehensive scene changes while maintaining consistent motion, \methodname{} successfully handles such complex transformations through its attention-based mechanism. Our qualitative and quantitative experiments demonstrate that \methodname{} significantly outperforms existing models in both fidelity and versatility even during drastic scene alterations.

\end{abstract}
    
\section{Introduction}
\label{sec:intro}

Recent advances in diffusion models demonstrated significant capabilities in generating high-quality images and videos. The emergence of text-to-video (T2V) generation \cite{guo2023animatediff, make-a-video, wang2023modelscope, yu2024efficient, chen2023control, hong2022cogvideo, yang2024cogvideox, zhou2022magicvideo, openai2024sora} models has opened new possibilities in creative content creation, enabling the synthesis of complex video sequences from user-provided text prompts. These models have shown a remarkable ability to generate diverse and visually compelling video content, marking a significant milestone in generative AI. 

Despite their success, current T2V models offer limited controllability, particularly in manipulating motion patterns. The ability to control motion in video generation is an important task for various creative applications.  Imagine a filmmaker in the early stages of planning a new movie scene, eager to explore various motion styles before committing to the labor-intensive process of shooting or animating the actual footage. Using \texttt{MotionFlow}, this filmmaker can repurpose video clips—such as a scene of a dog jumping into the lake (see Fig. \ref{fig:teaser})—and transfer these motions directly into the settings they envision. This capability allows the filmmaker to quickly prototype various motion effects and see how they might look with different characters or within different narrative contexts. For example, using the same motion from the dog video, they could experiment with \textit{a rabbit jumping into the river surrounded by blooming flowers} (see Fig. \ref{fig:teaser}). By enabling rapid experimentation with different motion dynamics, \texttt{MotionFlow} helps the filmmaker brainstorm, iterate on creative ideas, and overcome traditional time and resource constraints.

However, existing motion transfer methods face several limitations. They often struggle to balance motion fidelity and diversity, leading to issues like unwanted transfer of appearance and scene layout from the source video \cite{yatim2024space, zhao2025motiondirector}. Additionally, many approaches require extensive training \cite{tuneavideo, wang2024boximator, li2024animate} or fine-tuning on specific motion patterns  \cite{zhao2025motiondirector, ren2024customize, zhang2023motioncrafter, wang2024motion}, making them impractical for real-world applications where flexibility and efficiency are important. These limitations highlight the need for more sophisticated and practical solutions to motion transfer in video generation.

To address these challenges, we propose \methodname{}, a novel test-time approach that leverages the inherent capabilities of pre-trained video diffusion models without requiring additional training. While other approaches primarily rely on temporal attention features ~\cite{wang2024motion, bai2024uniedit, ren2024customize}, our method primarily leverages cross-attention features from existing videos to guide motion transfer. This approach enables the effective capture and transfer of motion information while remaining independent of the source video's appearance and scene composition. By visualizing cross-attention maps during both inversion and generation, we illustrate how linguistic elements influence object generation and motion (see Fig. \ref{fig:motivation}). These visualizations demonstrate how \methodname{} transfers motion dynamics by aligning the attention maps of the generated subject with those of the original, preserving motion patterns while adhering to the new edit prompt. Our contributions include:

\begin{itemize}
    \item We introduce the first test-time motion transfer method \textit{that leverages cross-attention maps} from pre-trained video diffusion models, eliminating the need for additional training, fine-tuning, or extra conditions.
    \item We provide comprehensive experimental results showing the effectiveness of our method across various scenarios and motion types.    Our approach achieves a balance between motion fidelity and diversity, generating the intended appearance and scene layout of the target video while accurately transferring motion patterns.
    \item We make our source code publicly available to enable further research and applications in this domain.
\end{itemize}

\section{Related Work}
\label{sec:relatedwork}

\subsection{Text-to-Video Diffusion Models}

Building on the success of diffusion-based Text-to-Image (T2I) models \cite{rombach2022high, imagen, dalle2, nichol2021glide}, Text-to-Video (T2V) generation models have made remarkable progress \cite{guo2023animatediff, make-a-video, wang2023modelscope, yu2024efficient, chen2023control, zhou2022magicvideo}. These models extend 2D diffusion frameworks by incorporating a temporal dimension, enabling the modeling of cross-frame dependencies and thus facilitating coherent video generation from text prompts.

\begin{figure}[t!]
    \centering
    \includegraphics[width=\linewidth]{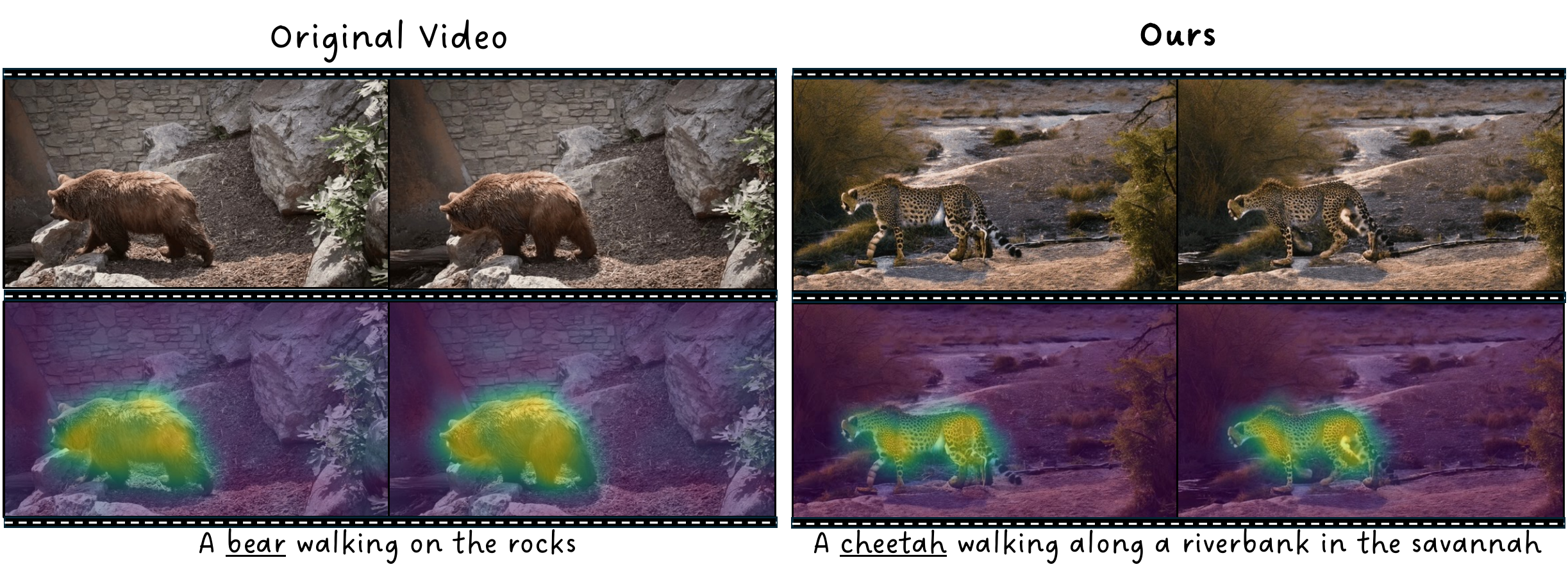}
    \vspace{-20pt}
    \caption{\textbf{Motivation.} Visualization of cross-attention maps for the subject tokens, showing how \methodname{} captures and transfers motion dynamics from the original video, ensuring accurate subject motion while adhering to new edit prompts.}
    \label{fig:motivation}
    \vspace{-15pt}
\end{figure}
Commonly, T2V models augment 2D diffusion architectures with temporal layers to explicitly model the relationships between video frames. For instance, AnimateDiff \cite{guo2023animatediff} and ModelScope \cite{wang2023modelscope} enhance pre-trained diffusion models by adding temporal layers and fine-tuning them for video generation. InstructVideo \cite{yuan2024instructvideo} takes a different approach by leveraging human feedback to refine video quality. Additionally, several works \cite{zhang2023controlvideo, yin2023dragnuwa, lin2023videodirectorgpt, lian2023llm, chen2023motion} introduce conditioning inputs—such as depth maps, bounding boxes, and motion trajectories—to allow for more precise control over object shapes and movements within generated videos.

 \subsection{Attention-Based Guidance}
Attention-based guidance has emerged as a key technique for improving the quality and controllability of T2I generation. Methods such as Attend-and-Excite \cite{chefer2023attend}, and CONFORM \cite{meral2023conform} apply attention constraints to optimize latent features during inference, addressing issues like subject omission and incorrect attribute binding, while methods like CLoRA \cite{meral2024clora} and Bounded Attention \cite{dahary2024yourself} further enhance multi-subject generation by managing cross-attention in complex compositions. These advances in attention-based guidance for T2I generation provide a foundation for more sophisticated control in T2V models, where temporal consistency is crucial.

\subsection{Video Motion Editing}

While Text-to-Video (T2V) models are designed to control motion through text prompts, they often struggle with complex or nuanced motions. To address this, recent methods have introduced bounding boxes for more precise control, either during training \cite{li2024animate, wang2024boximator} or at inference time \cite{jain2024peekaboo, ma2023trailblazer}.

Another line of work focuses on transferring motion from a reference video. Fine-tuning approaches store motion in the model's weights, while inversion-based methods store motion in model features. For example, Tune-a-Video \cite{tuneavideo} adapts text-to-image models by adding spatiotemporal attention layers, training only the motion-specific components. Similarly, MotionDirector \cite{zhao2025motiondirector} separates motion and appearance using a dual-path LoRA architecture. Other methods, such as DreamVideo \cite{wei2024dreamvideo} and Customize-A-Video \cite{ren2024customize}, use separate branches for appearance and motion learning. Wang et al. \cite{wang2024motion} learn motion embeddings by training over the original video using temporal attention layers.

Inversion-based editing methods, initially developed for image editing \cite{hertz2022prompt, tumanyan2023plug}, have also been adapted for video. DDIM inversion \cite{ddim} enables reconstruction through backward diffusion, as used in methods like DMT \cite{yatim2024space}, UniEdit \cite{bai2024uniedit} and VMC \cite{jeong2024vmc}. DMT uses a space-time feature loss that leverages DDIM inversion and UNet activations. UniEdit and VMC blend fine-tuning with inversion to adjust temporal layers while maintaining content fidelity. However, a common limitation is the assumption that the features of the reference and target videos are identical, which can pose challenges when generating videos with different geometries.

In contrast, our proposed method, \methodname{}, introduces a novel test-time approach that leverages cross-attention features from pre-trained video diffusion models. This allows for effective motion transfer without additional training or fine-tuning, overcoming the limitations of existing methods. By capturing and transferring motion independently of the source video's appearance and scene composition, \methodname{} achieves a balance between motion fidelity and diversity, offering enhanced flexibility and control in video motion transfer tasks.

\section{Methodology}
\label{sec:Methodology}

\subsection{Diffusion Models}

Diffusion models iteratively transform input noise $x_T \sim \mathcal{N}(0, \mathbf{I})$ into a meaningful sample $x_0$ through a structured denoising process. Denoising Diffusion Implicit Models (DDIM) \cite{ddim} enable this process deterministically, converting $x_T$ into a clear sample $x_0$. The reverse process, known as inversion, reconstructs the initial noise $x_T$ responsible for generating $x_0$ and the entire sequence of noisy latents $\{x_t\}_{t=1}^T$.

Latent diffusion models, like Stable Diffusion \cite{rombach2022high}, operate within the latent space of an autoencoder. The input image $x$ is compressed into a lower-dimensional latent representation $z = \mathcal{E}(x)$ via an encoder $\mathcal{E}$, and then reconstructed by a decoder $\mathcal{D}$. For instance, Stable Diffusion encodes an image $x \in \mathbb{R}^{H \times W \times C}$ into a latent space $z \in \mathbb{R}^{h \times w \times d}$, where $H \gg h$, $W \gg w$, and $d$ is the latent dimension, typically $d=4$.

Extending this approach to video data, latent video diffusion models encode video input $x\in\mathbb{R}^{F \times H \times W \times C}$ into a latent space $z \in \mathbb{R}^{f \times h \times w \times d}$, where $f \leq F$. These models enhance existing T2I frameworks by incorporating temporal layers, which integrate convolution and attention mechanisms, and are fine-tuned on video datasets \cite{blattmann2023align, wang2023modelscope}. In our study, we use the publicly available ZeroScope T2V model \cite{cerspense2023zeroscope}, which extends Stable Diffusion by integrating temporal convolution and attention layers.

\subsection{Attention Mechanisms}

The backbone of our method, ZeroScope, built on a UNet architecture, integrates two key components for influencing video motion: temporal attention layers in the temporal module and cross-attention layers in the spatial module.
Cross-attention layers incorporate text-based information, allowing the text prompt to guide the content and structure of the generated video. This ensures that linguistic elements, such as nouns and verbs, are accurately translated into visual features, guiding object generation and their associated motions. Temporal attention layers establish inter-frame connections, ensuring smooth and coherent motion across frames, while self-attention within spatial blocks preserves spatial consistency.

The query features, $Q \in \mathbb{R}^{F \times N \times D}$, represent the latent features of the model, where $N = h \times w$ (with $h \ll H$ and $w \ll W$) is the spatial resolution in the latent space, and $D$ is the feature dimension. The key features, $K \in \mathbb{R}^{L \times D}$, are derived from the text encoder (e.g. CLIP \cite{radford2021learning}), where $L$ represents the number of tokens in the input text prompt. Consequently, the cross-attention map has dimensions $F \times N \times L$, where $F$ is the number of frames, $N = h \times w$ denotes the spatial resolution, and $L$ is the number of tokens from the text prompt. The attention map at time $t$ is calculated as $A^{t} = \text{Softmax}(Q K^\intercal / \sqrt{d})$ where $t$ is the timestep and $d$ is the dimension of the keys.

\subsection{Our Method}

\begin{figure*}
    \centering
    \includegraphics[width=1\linewidth]{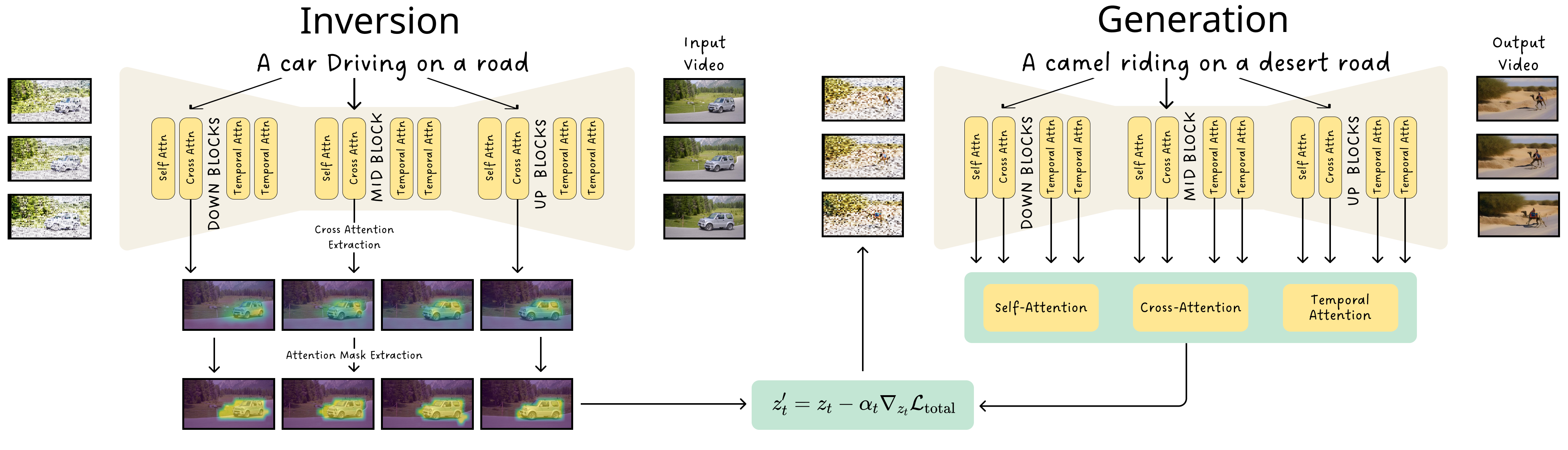}
    \vspace{-20pt}
    \caption{\textbf{Overview of \methodname{} framework.} Our invert-then-generate method operates in two main stages: (1) \textbf{Inversion}, where DDIM inversion is used to extract latent representations and cross-attention maps from the original video, generating target masks that capture the subject's motion and spatial details; (2) \textbf{Generation}, where these masks and a text prompt guide the creation of a new video, aligning with the original video's motion dynamics and spatial layout while adhering to the semantic content of the prompt.}
    \label{fig:framework}
    \vspace{-15pt}
\end{figure*}

We propose a framework for generating a new video $\hat{V}$ by transferring the motion dynamics of a specific subject from an original video $V$, while ensuring compliance with a target text prompt $P$. Our method leverages ZeroScope, a pre-trained latent text-to-video (T2V) diffusion model, and involves two key steps: (1) DDIM inversion of the original video to extract latent representations and cross-attention-based extraction of subject-specific motion, and (2) guided video generation using the extracted motion and the target text prompt.

\noindent\textbf{Inversion.} We begin by encoding the original video $V \in \mathbb{R}^{F \times H \times W \times C}$, where $F$ is the number of frames, $H$ is the height, $W$ is the width, and $C$ is the number of channels, into a lower-dimensional latent space using the encoder. The Variational Autoencoder   $\mathcal{E}$ takes each frame $v_f \in \mathbb{R}^{H \times W \times C}$ and encodes it into a latent representation $z_f \in \mathbb{R}^{h \times w \times d}$, where $h \ll H$, $w \ll W$, and $d$ is the latent dimension. The noisy latents from DDIM inversion are then used to extract cross-attention maps, which provide crucial spatial and temporal information for guiding the new video generation. Specifically, the DDIM inversion process generates a sequence of noisy latents $\{z_t^f\}_{t=1}^T$ for each frame, where $T$ is the total number of timesteps in the diffusion process.

\noindent\textbf{Attention Extraction.} To capture subject-specific motion, we extract cross-attention maps from the noisy latent representations. Given a text prompt $P = \{p_1, p_2, \ldots, p_L\}$, we define key token sets as $S^{*} = \{s_0, s_1, \ldots, s_l\}, l\le L$, which describe the subject and action.

During DDIM inversion of the original video, cross-attention maps $A_{s,f}^{t}$ for token $s$ at timestep $t$ and frame $f$ are extracted for key tokens across frames, providing crucial spatial and temporal insights into the subject's position and motion dynamics, as shown in  Fig. \ref{fig:motivation} and Fig. \ref{fig:framework}.

In our experiments with \methodname{}, we focus on leveraging specific attention layers from the UNet architecture to optimize motion transfer. We extract attention from the middle block, the last block of the down-sampling layers, and the first block of the up-sampling layers. Previous studies \cite{meral2023conform, hertz2022prompt} have suggested that bottleneck layers are more expressive in terms of spatial information. However, relying solely on the bottleneck layer restricts the resolution of the cross-attention maps to $\{9 \times 5\}$ for generating videos with a resolution of $\{576 \times 320\}$. To address this, we incorporate the last down-sampling block and the first up-sampling block, both of which provide a higher resolution of $\{18 \times 10\}$, ensuring more detailed spatial information.

The cross-attention maps, along with the initial noisy latent representations, guide the generation of the new video $\hat{V}$.  To implement this guidance, we convert the attention maps $A_{s,f}^{t}$ into binary masks $M_{s,f}^{t}$, using an adaptive thresholding formula \cite{tang2022daam} for each cross-attention map corresponding to the key tokens at each frame:
\vspace{-5pt}
\begin{equation}
    M^{t}_{s,f}[x,y] = \mathbb{I} \left( A^{t}_{s,f}[x,y] > \tau \max_{i,j} A^{t}_{s,f}[i,j] \right)
    \label{eq:mask}
\end{equation}

Here, $\mathbb{I}$ is the indicator function, and $\tau$ is a threshold parameter that determines the significance of the attention values based on the maximum attention weight. These binary masks ensure that the motion and actions in the generated video align with the original cross-attention maps.

By controlling the spatial position and trajectory of cross-attention maps, our method offers fine-grained control over object motion and behavior, ensuring precise alignment between the text prompt and the generated video.

\noindent\textbf{Guided Generation.} To align the generated video with the original video's motion dynamics and spatial-temporal characteristics $V$, we optimize the noisy latent representations through backpropagation. This is achieved by applying a series of loss functions to the cross-attention, self-attention, and temporal attention layers, guiding the latent signal toward the desired motion and behavior.

The cross-attention loss ensures that the attention maps for the key tokens (e.g. subject and action) in the generated video $\hat{V}$ align with those from the original video $V$, preserving the subject’s motion and behavior by maximizing the attention within the binary mask $M_{s,f}^{t}$:
\vspace{-5pt}
\begin{equation} 
\mathcal{L}_{s,f} = 1 - \frac{M_{s,f} \cdot A_{s,f}}{A_{s,f}} 
\label{eq:loss}
\end{equation}

The self-attention loss operates similarly to the cross-attention loss but targets the self-attention layers, which capture relationships between different spatial locations within the same frame. This loss ensures spatial consistency for the subject within the frames of the generated video.

The temporal attention loss preserves the temporal dynamics of the subject and action tokens are preserved across frames. Temporal attention layers capture relationships between frames, ensuring smooth and coherent motion. This loss aligns the temporal attention maps in the generated video with those from the original.

All three losses—cross-attention, self-attention, and temporal attention—work together to guide the generation process. The total loss function is a weighted combination of these three components:
\vspace{-5pt}
\begin{equation} 
    \mathcal{L}_{\text{total}} = \lambda_{\text{cross}} \mathcal{L}_{\text{cross}} + \lambda_{\text{self}} \mathcal{L}_{\text{self}} + \lambda_{\text{temporal}} \mathcal{L}_{\text{temporal}} 
\label{eq:total_loss}
\end{equation}
Here, $\lambda_{\text{cross}}$, $\lambda_{\text{self}}$, $\lambda_{\text{temporal}}$ are the weights for each corresponding loss.

We optimize this total loss using gradient descent, updating the latent representations at each timestep $t$ as follows:
\vspace{-10pt}
\begin{equation} z^\prime_t = z_t - \alpha_t \nabla_{z_t} \mathcal{L}_{\text{total}} \label{eq:optimization} \end{equation}

Here, $\alpha_t$ is the learning rate, and $\nabla_{z_t} \mathcal{L}_{\text{total}}$ represents the gradient of the total loss with respect to the latent representation at timestep $t$. Through iterative backpropagation, we refine the latent representations to ensure that the generated video $\hat{V}$ respects the desired motion dynamics and behavior while maintaining the spatio-temporal characteristics of the original video $V$. Our framework is summarized in Alg. \ref{alg:guided_generation} and Fig. \ref{fig:framework}.

\begin{algorithm}
\caption{MotionFlow: Attention-Driven Motion Transfer in Video Diffusion Models}
\label{alg:guided_generation}
\KwIn{Original video $V$, text prompt $P$, pre-trained T2V model, total timesteps $T$}
\KwOut{Generated video $\hat{V}$}

\textbf{Step 1: Encode Original Video}\\
\ForEach{frame $v_f \in V$}{
    Encode $v_f$ into latent representation $z_f$
}

\textbf{Step 2: Apply DDIM Inversion}\\
\ForEach{timestep $t$}{
    Generate noisy latents $z^t$ using DDIM inversion
}

\textbf{Step 3: Extract Cross-Attention Maps}\\
\ForEach{timestep $t$ and frame $f$ and token $s$}{
    Extract cross-attention maps $A_{s,f}^{t}$
}

\textbf{Step 4: Convert to Binary Masks}\\
\ForEach{attention map $A_{s,f}^{t}$}{
    Convert $A_{s,f}^{t}$ to binary mask $M_{s,f}^{t}$
}

\textbf{Step 5: Guided Video Generation}\\
\ForEach{timestep $t$ and frame $f$ and token $s$}{
    Denoise $\hat{z}^t$ guided by $M_{s,f}^{t}$\\
    Compute attention-based losses\\
    Update latent $\hat{z}^t$ using gradient descent
}

\textbf{Step 6: Output Generated Video}\\
\Return{$\hat{V}$}
\end{algorithm}

\section{Experiments}
\label{sec:exp} 

\noindent \textbf{Experimental Setup.} For the inversion of the original video, we use DDIM inversion to obtain the initial noise latent. During the generation process, we set the threshold parameter $\tau = 0.4$ in Eq. \ref{eq:mask} to determine the significance of attention values. For optimization, we utilize a learning rate $\alpha = 5.0$ in Eq. \ref{eq:optimization}, and assign equal weights to the loss functions, with $\lambda_\text{cross} = \lambda_\text{self} = \lambda_\text{temporal} = 1.0$.

The generation process involves performing latent updates for 20 steps out of a total of 50 backward diffusion steps, similar to \cite{yatim2024space}, with 20 iterations per update. To mitigate the introduction of unwanted artifacts in the output, we halt optimization after the 20th step, allowing the remaining steps to proceed without further updates.

For our experiments, we selected a range of videos from the DAVIS~\cite{davis} dataset, which is widely used in video editing and motion transfer research. This dataset provides a robust foundation for evaluating the effectiveness of \methodname{} in diverse motion transfer scenarios.\\

\begin{figure*}
    \centering
    \includegraphics[width=0.9\linewidth]{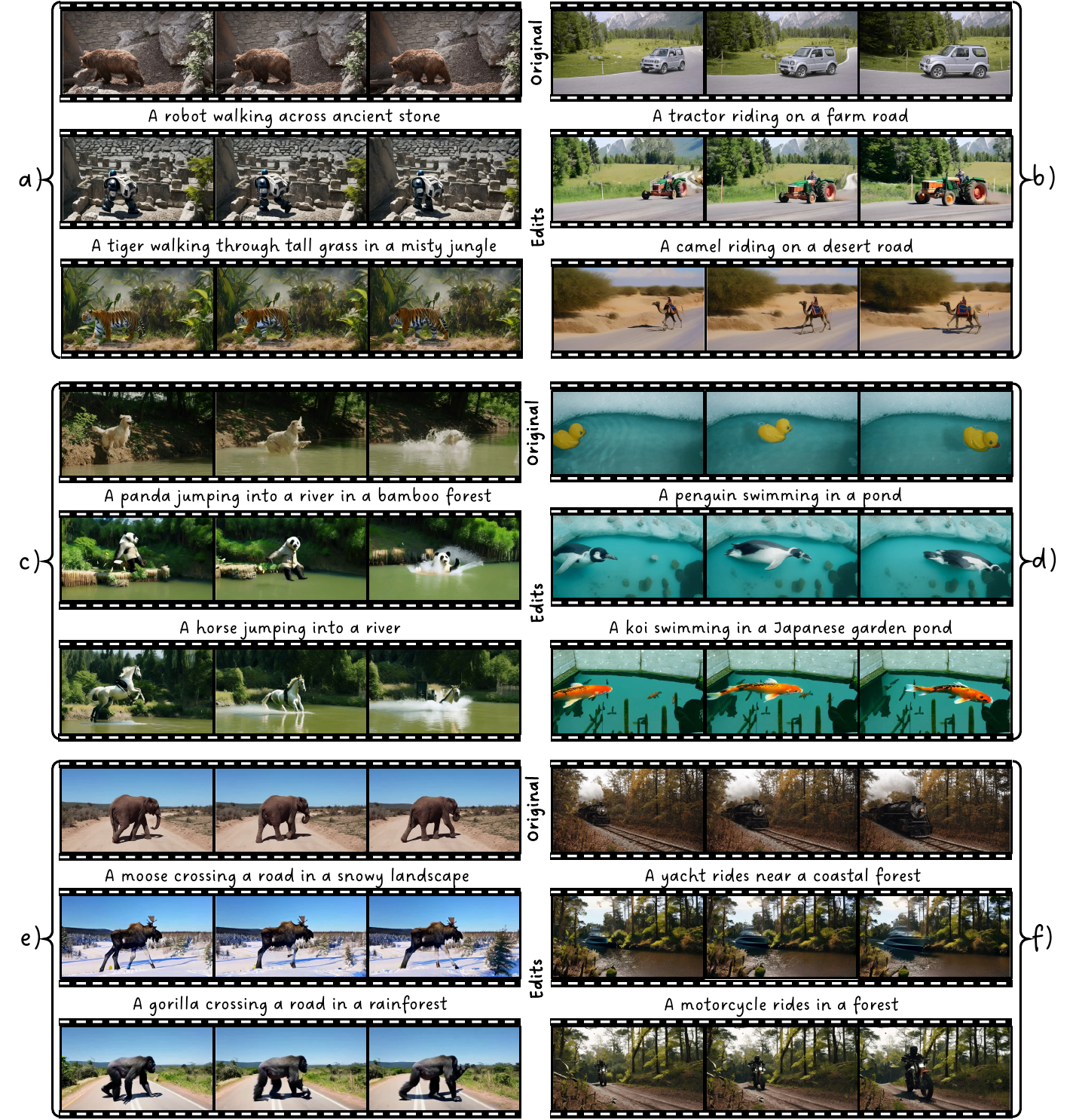}
    \vspace{-5pt}
    \caption{\textbf{Qualitative Results.}  \methodname{} can successfully transfer a wide variety of motion types, ranging from single to multiple motions and from simple to complex motion patterns. Additionally, it can either maintain the original scene layout or significantly alter it based on the user-provided text prompt. Please refer to the supplementary material where the actual videos are provided. }
    \label{fig:qual}
    \vspace{-15pt}
\end{figure*}

\noindent \textbf{Baselines and Metrics.} We evaluate \methodname{} against state-of-the-art video motion transfer methods: DMT \cite{yatim2024space}, VMC \cite{jeong2024vmc}, Motion Director \cite{zhao2025motiondirector}, and Motion Inversion \cite{wang2024motion}, using 100 randomly selected video-prompt pairs for each method. To assess performance, we employ three metrics: Motion Fidelity Score \cite{yatim2024space}, which measures how well the generated video preserves the motion patterns from the source video, evaluating the accuracy of motion dynamics such as speed, direction, and style; Temporal Consistency, which assesses the smoothness and coherence of motion across frames by calculating the average cosine similarity between CLIP image features of consecutive frames, ensuring stable transitions and natural motion flow; and Text Similarity, which evaluates how accurately the generated video aligns with the input text prompt by computing the cosine similarity between CLIP embeddings of the video frames and the prompt, ensuring that the visual content reflects the intended modifications.

\subsection{Qualitative Experiments}

\begin{figure*}[t]
    \centering
    \includegraphics[width=1\linewidth]{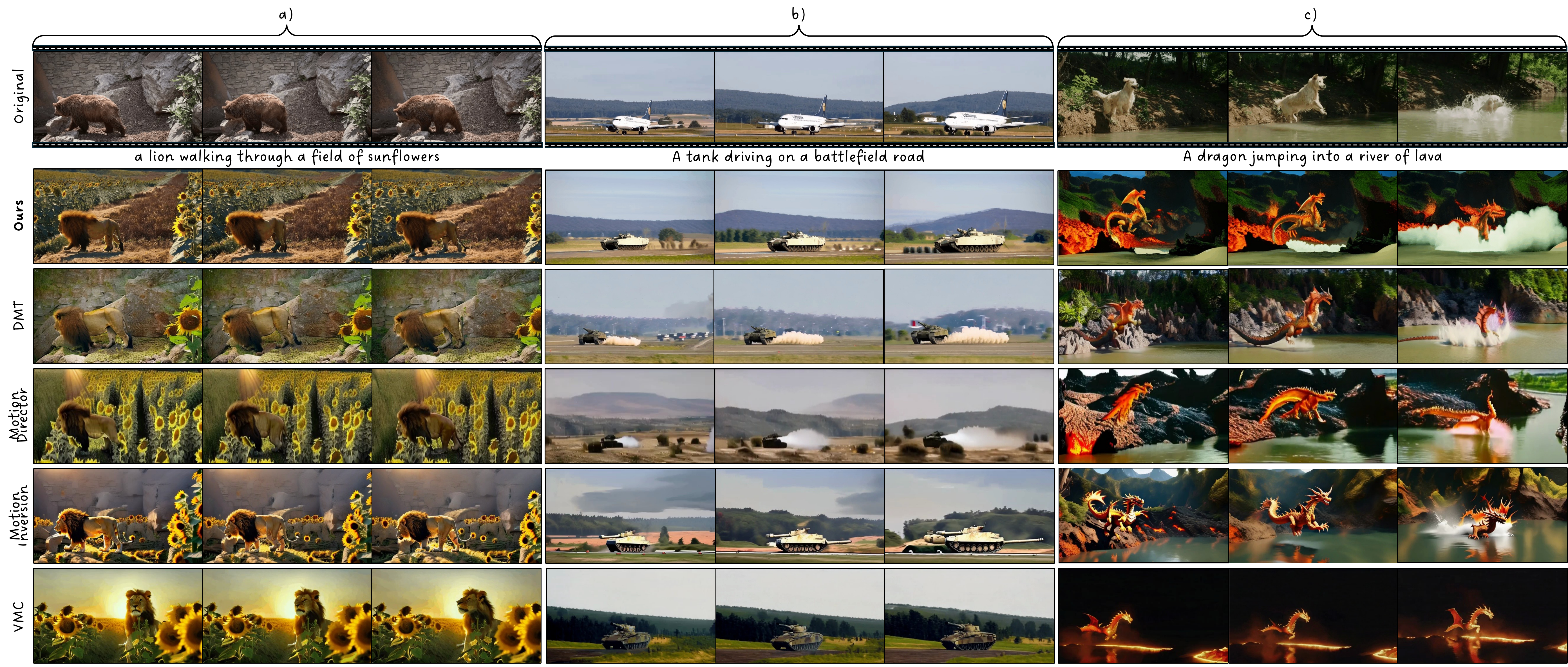}
    \vspace{-15pt}
    \caption{\textbf{Comparison.} Qualitative comparison of our method, \methodname{}, with DMT \cite{yatim2024space}, MotionDirector \cite{zhao2025motiondirector}, Motion Inversion \cite{wang2024motion} and VMC \cite{jeong2024vmc}}
    \label{fig:comparison}
    \vspace{-20pt}
\end{figure*}

As shown in Fig. \ref{fig:qual}, \methodname{} demonstrates remarkable flexibility in motion transfer across diverse scenarios while offering control over scene composition. In Fig. \ref{fig:qual} (a), our method successfully transfers motion between significantly different animals and objects, mapping a bear's movement to both a robot and a tiger. This highlights \methodname{}'s ability to generate entirely new scene layouts with different subjects, overcoming the common limitation of being constrained by the source video's background. Conversely, our method can also preserve the original scene layout, as seen in Fig. \ref{fig:qual} (e), where we transfer motion from an elephant to a moose and a gorilla while maintaining the original background composition. Moreover, our method excels at transferring motion between fundamentally different object categories, as shown in Fig. \ref{fig:qual} (f) where we successfully map the motion of a train to a motorbike while preserving the distinctive movement patterns. These results underscore \methodname{}'s versatility in handling diverse motion transfer scenarios and its ability to either maintain or alter scene layouts as needed, offering high flexibility in video motion transfer applications. Full videos are available in the Supplementary Material.

\noindent \textbf{Qualitative Comparison.} In Fig. \ref{fig:comparison}, \methodname{}'s motion transfer capabilities are compared across different videos against benchmark methods. In Fig. \ref{fig:comparison} (a), \methodname{} successfully transfers motion from the subject of the original video and generates a background aligned with the prompt, unlike DMT and Motion Inversion. While both MotionDirector and VMC generate videos aligned with the edit prompt, MotionDirector's output suffers from poor quality, and VMC produces a video with low motion fidelity to the original. In Fig. \ref{fig:comparison} (c), all methods except VMC, which generally exhibits low motion fidelity, were able to transfer motion. However, only our method successfully generates a scene aligned with the edit  `river of lava'.

\subsection{Quantitative Experiments}

\begin{figure}
    \centering
    \includegraphics[width=0.8\linewidth]{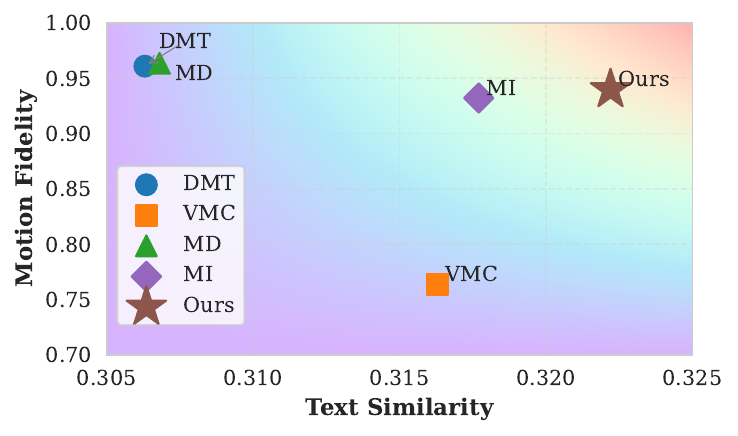}
    \vspace{-15pt}
    \caption{\textbf{Evaluation.} CLIP text similarity versus Motion Fidelity scores for each baseline. Our method exhibits a better balance between these two metrics.}
    \label{fig:comparison-plot}
    \vspace{-15pt}
\end{figure}

\begin{table*}[t]
\centering
\renewcommand{\arraystretch}{1}
\setlength{\tabcolsep}{6pt}
\begin{tabular}{c|ccc|ccc}
\hline
\multirow{2}{*}{Methods} & \multicolumn{3}{c|}{Quantitative Results}           & \multicolumn{3}{c}{User Study}                \\ \cline{2-7} 
 &
  \begin{tabular}[c]{@{}c@{}}Text\\ Similarity\end{tabular} &
  \begin{tabular}[c]{@{}c@{}}Motion\\ Fidelity\end{tabular} &
  \begin{tabular}[c]{@{}c@{}}Temporal\\ Consistency\end{tabular} &
  \begin{tabular}[c]{@{}c@{}}Text\\ Alignment\end{tabular} &
  \begin{tabular}[c]{@{}c@{}}Motion\\ Alignment\end{tabular} &
  \begin{tabular}[c]{@{}c@{}}Motion\\ Smoothness\end{tabular} \\ \hline
DMT \cite{yatim2024space}                      & 0.3063         & 0.960          & 0.934          & 0.20          & 0.17          & 0.19          \\
MD  \cite{zhao2025motiondirector}                      & 0.3068          & \textbf{0.963} & 0.928          & 0.10          & 0.12          & 0.11          \\
VMC \cite{jeong2024vmc}                     & 0.306          & 0.763          & \textbf{0.961} & 0.15          & 0.15          & 0.16          \\
MI  \cite{wang2024motion}                     & 0.317          & 0.930          & 0.941          & 0.13          & 0.13          & 0.15          \\ \hline
\textbf{Ours}            & \textbf{0.322} & 0.940          & 0.941          & \textbf{0.42} & \textbf{0.43} & \textbf{0.39} \\ \hline
\end{tabular}
\vspace{-8pt}
    \caption{\textbf{Quantitative Comparisons and User Study} Quantitative comparisons for Text Similarity, Motion Fidelity, and Temporal Consistency Scores. User preferences for text alignment to edit prompt, motion alignment to the original video, and motion smoothness of the generated video for DMT, MotionDirector, VMC, Motion Inversion, and \methodname{}. }
    \label{tab:comparison}
    \vspace{-15pt}
\end{table*}

Table \ref{tab:comparison} and Fig. \ref{fig:comparison-plot} present the performance of our approach across key metrics, highlighting its advantages over existing methods. Our method consistently outperforms baselines in text similarity, demonstrating better alignment with target prompts while retaining strong motion fidelity and temporal consistency.

This improved performance can be attributed to three key factors: (1) By leveraging cross-attention maps, our framework achieves precise motion transfer, preserving the integrity of original motion patterns without requiring fine-tuning. This sets it apart from training-intensive methods like DMT and MotionDirector. (2) As shown in Fig. \ref{fig:comparison-plot}, our method balances text similarity and motion fidelity, enabling accurate motion adaptation to prompt specifications. High motion fidelity alone can sometimes penalize necessary edits, but our method strikes an effective compromise, maintaining fidelity to the original while allowing flexibility for editing tasks. (3) Our approach closely aligns with the target prompt, resulting in higher text similarity scores and improved semantic coherence in generated videos. This alignment surpasses methods such as VMC, which struggles with motion fidelity, and Motion Inversion, although it uses an alternative diffusion model backbone (MotionCraftV2 \cite{zhang2023motioncrafter}).

In summary, our framework achieves a strong balance across metrics. It provides competitive motion fidelity and temporal consistency while maintaining superior adherence to user-defined prompts, as evidenced by its higher text similarity scores.

\noindent\textbf{User Study.} To evaluate the perceptual quality of our motion transfer results, we conducted a user study on Amazon Mechanical Turk with 50 participants. Each participant viewed 30 sets of videos, each set containing five generated videos: one from our method, four from baseline methods (Motion Inversion, DMT, VMC, and Motion Director), along with the original video and the corresponding edit prompt. Participants were asked to select the best video based on three key criteria: Motion Fidelity (how well the original motion was preserved), Visual Quality (overall appearance and coherence), and Prompt Alignment (how accurately the video matched the edit prompt). As shown in Table \ref{tab:comparison}, our method consistently ranked higher across all criteria according to user preferences.

\noindent\textbf{Processing Times.} We compared the processing times for each method using Nvidia L40 GPUs. Each method involves an initial setup phase (training, fine-tuning, or inversion) followed by video generation, requiring 49 seconds for inversion and 376 seconds for generation, totaling 425 seconds.  DMT takes 258 seconds for inversion and 332 seconds for generation (590 seconds total), Motion Director requires 410 seconds for fine-tuning and 67 seconds for generation (477 seconds total), VMC takes 227 seconds for training and 503 seconds for generation (730 seconds total), and Motion Inversion requires 195 seconds for training and 30 seconds for generation, totaling 225 seconds.

\begin{figure} 
    \centering
    \includegraphics[width=\linewidth]{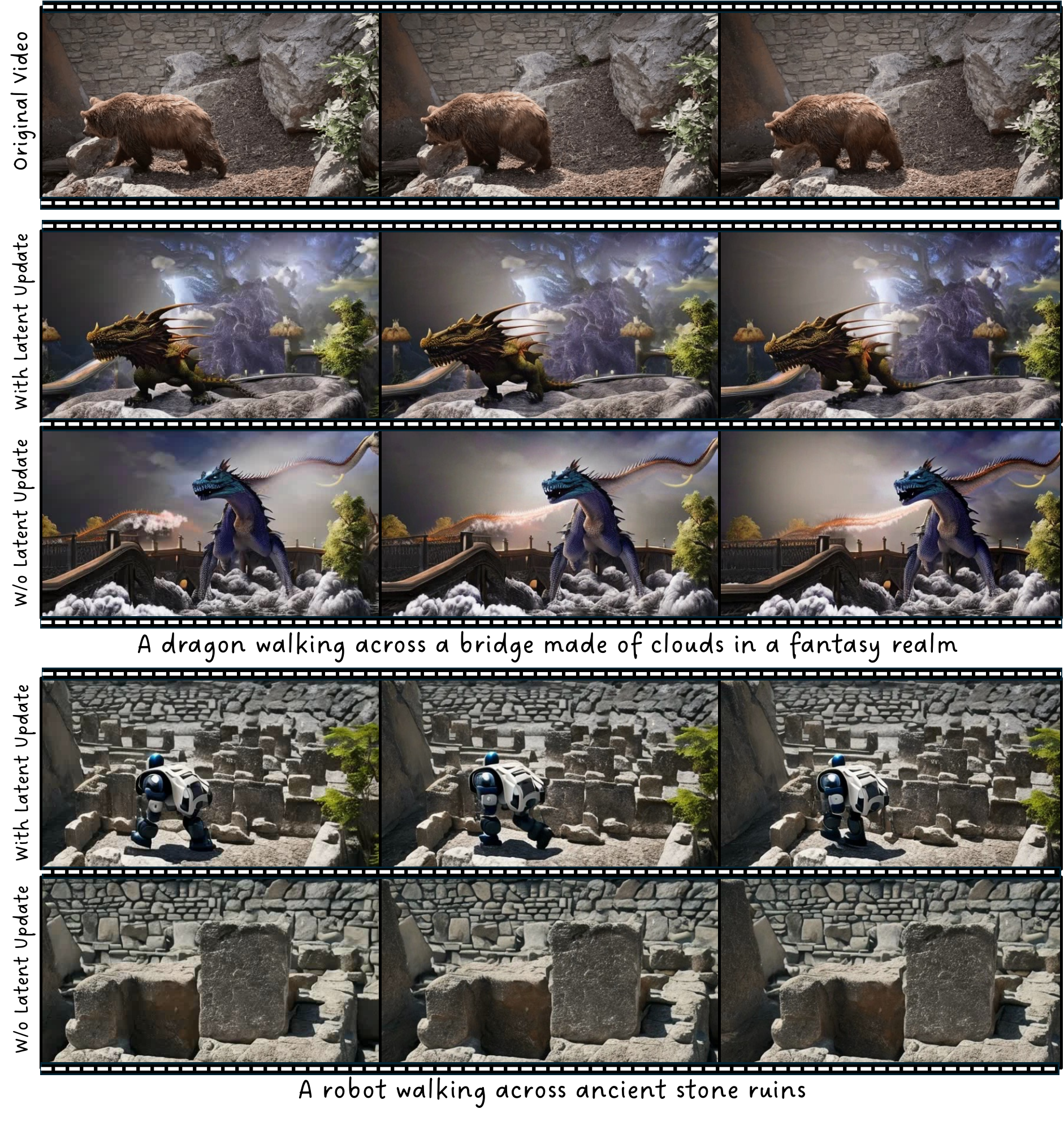} 
    \vspace{-20pt}
    \caption{Ablation study on latent updates. Without latent updates guided by cross-attention, inverted latents may fail to preserve motion or generate the intended subject. Please see Supplementary Material for full videos.} 
    \label{fig:ablation} 
    \vspace{-15pt}
\end{figure}

\noindent \textbf{Ablation Study.} To evaluate the importance of latent updates in motion transfer, we conducted an ablation study (Fig. \ref{fig:ablation}) by omitting the latent update step and using only the initial noise latent from DDIM inversion. The results show that while DDIM inversion provides a high-level structure and preserves camera motion, it often fails to capture detailed subject motion. For example, in the prompt `A dragon walking across a bridge made of clouds in a fantasy realm', The results show that while DDIM inversion provides a high-level structure and preserves camera motion, it often fails to capture detailed subject motion. For example, in the prompt `A robot walking across ancient stone ruins', the model struggles to generate the intended subject. In contrast, our method, which incorporates cross-attention-guided latent updates, accurately captures both the subject and its motion, ensuring correct spatial placement. This ablation study demonstrates the critical role of latent updates in achieving precise motion transfer and subject generation. Please see Supplementary Material for full videos.

\section{Limitation and Societal Impact}

While \methodname{} demonstrates strong performance in motion transfer tasks, we acknowledge some limitations that present opportunities for future research. Our method's fundamental reliance on attention maps from pre-trained video diffusion models makes it sensitive to the quality  of these underlying models. When attention maps are noisy or improperly focused due to the pre-trained model's limitations, our motion transfer quality may degrade accordingly. While \methodname{} has the potential to significantly impact creative industries, it is essential to consider ethical implications, such as the misuse of the technology for creating deceptive   content. Ensuring responsible use through clear guidelines and safeguards will help maximize its positive societal contributions while minimizing potential risks.

\section{Conclusion}

In this paper, we introduced \methodname{}, a novel approach to video motion transfer that leverages cross-attention maps from pre-trained video diffusion models without requiring additional training. Our method addresses a significant challenge in video generation by enabling precise control over motion patterns while maintaining the flexibility to either preserve or modify scene compositions as desired. Through comprehensive experiments, we demonstrated that \methodname{} achieves state-of-the-art performance across various metrics, successfully handling diverse scenarios from simple object transformations to complex cross-category motion transfers. The method's ability to work with drastically different objects (e.g., train to motorbike) and animals (e.g., bear to elephant) while offering control over scene preservation demonstrates its versatility and practical utility. By making our code public, we hope to facilitate further research in this direction and enable practical applications in content creation, animation, and video editing.
\vspace{-20pt}
{
    \small
    \bibliographystyle{ieeenat_fullname}
    \bibliography{main}
}
\clearpage
\setcounter{section}{0}  %
\renewcommand{\thesection}{\Alph{section}}  %
\maketitlesupplementary

\section{User Study}
\label{sec:user-study}

\begin{figure}[h]
    \centering
    \includegraphics[width=0.9\linewidth]{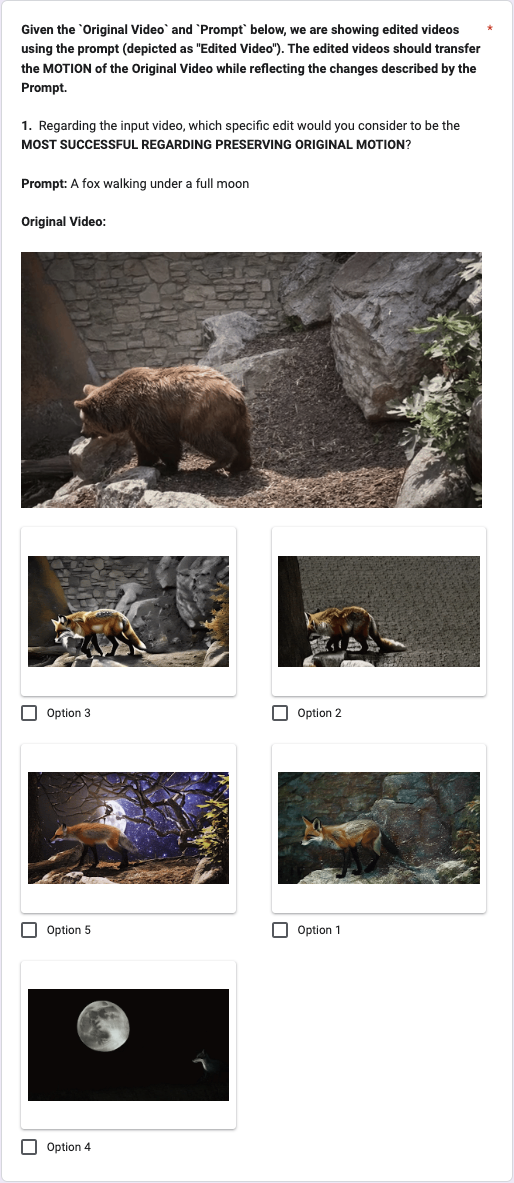}
    \caption{An example question used in the user study. Participants were asked to evaluate multiple videos based on motion fidelity, motion smoothness, and text fidelity.}
    \label{fig:user-study}
\end{figure}

An example question from our User Study is shown in Fig. \ref{fig:user-study}. Participants were asked to evaluate three aspects of the generated videos across 10 different edits. Each edit included 5 generated videos from our method and competitors \cite{yatim2024space, zhao2025motiondirector, wang2024motion, jeong2024vmc} alongside the original input video. The questions focus on:

\begin{itemize}
    \item  Motion Fidelity:  \textit{Regarding the input video, which specific edits would you consider to be the most successful regarding preserving original motion?}

    \item Motion Smoothness: \textit{Regarding the input video, which specific edit would you consider to be most successful regarding the smoothest motion? }

    \item Text Fidelity: \textit{Regarding the input video, which specific edit would you consider to the top video that aligns with prompt?}
\end{itemize}

Participants were required to select the most suitable video for each question based on their subjective judgment. The aggregated results, detailed in Section \ref{sec:user-study}, provide insights into MotionFlow's performance relative to competing methods.

\end{document}